\documentclass{article}

\usepackage{PRIMEarxiv}

\usepackage[utf8]{inputenc} 
\usepackage[T1]{fontenc}    
\usepackage{hyperref}       
\usepackage{url}            
\usepackage{booktabs}       
\usepackage{amsfonts}       
\usepackage{nicefrac}       
\usepackage{microtype}      
\usepackage{lipsum}
\newcommand{\JournalTitle}[1]{#1}
\usepackage{fancyhdr}       
\usepackage{graphicx}       
\graphicspath{{media/}}     
\usepackage{float}
\usepackage{listings}%
\usepackage{subfigure}
\usepackage{lmodern}
\usepackage{mathptmx}
\usepackage{titlesec}
\usepackage{amsmath} 
\titlespacing*{\section}{0pt}{0pt plus 1pt minus 1pt}{0pt plus 1pt minus 1pt}
\title{A Curated and Re-annotated Peripheral Blood Cell Dataset Integrating Four Public Resources

}

\author{
  Lu Gan \\
  Northern Arizona University \\
  Depaetment of Computer Information Technology\\
  Flagstaff, AZ,  USA\\
  \texttt{lg2465@nau.edu} \\
   \And
  Xi Li \\
  Independent Researcher \\
  Chengdu, China\\
  \texttt{reilixi723@gmail.com} \\
  \AND
  Xichun Wang \\
  Northern Arizona University \\
  Depaetment of Computer Information Technology\\
  Flagstaff, AZ,  USA\\
  \texttt{Steve.Wang@nau.edu} \\
}

\begin{document}
\maketitle

\begin{abstract}
We present TXL-PBC, a curated and re-annotated peripheral blood cell dataset constructed by integrating four publicly available resources: Blood Cell Count and Detection (BCCD), Blood Cell Detection Dataset (BCDD), Peripheral Blood Cells (PBC), and Raabin White Blood Cell (Raabin-WBC). Through rigorous sample selection, semi-automatic annotation using the YOLOv8n model, and comprehensive manual review, we ensured high annotation accuracy and consistency. The final dataset contains 1,260 images and 18,143 bounding box annotations for three major blood cell types: white blood cells (WBC), red blood cells (RBC), and platelets. We provide detailed visual analyses of the data distribution, demonstrating the diversity and balance of the dataset. To further validate the quality and utility of TXL-PBC, we trained several mainstream object detection models, including YOLOv5s, YOLOv8s, YOLOv11s, SSD300, Faster R-CNN, and RetinaNet, and report their baseline performance. The TXL-PBC dataset is openly available on Figshare and GitHub, offering a valuable resource for the development and benchmarking of blood cell detection models and related machine learning research.
\end{abstract}


\section{Introduction}
In clinical medical diagnosis, the analysis, detection, and counting of blood cells are important indicators for doctors to diagnose diseases. Nowadays, researchers have integrated artificial intelligence models into analysis, detection, and counting of blood cells\cite{mlbloodcell,mlbloodcell1}. To enhance the accuracy of AI model detection, it is crucial to strictly screen the quality and diversity of blood cells. However, many hospitals are reluctant to disclose their datasets due to patient privacy and security concerns, making blood cell datasets scarce\cite{dataprivacy,datapravicy1,datapravicy2}.

\hspace{2em}We found the public available Blood Cell Count and Detection (BCCD) dataset\cite{BCCD} on Github open-source website. However, this dataset contains only 364 images, and many of them have overlapping red blood cells, making it difficult to distinguish individual cells. We also observed annotation errors in the dataset, particularly frequent missed labels for red blood cells. These issues can negatively impact the performance of machine learning models. Therefore, we performed a thorough screening and re-annotation of this dataset to improve its quality and suitability for research. To further increase the diversity of our samples, we additionally incorporated three publicly available cell image datasets into our collection: the Blood Cell Detection Dataset\cite{BCD}, the Peripheral Blood Cells (PBC) dataset\cite{PBC}, and the Raabin-WBC dataset\cite{Raabin-WBC}. We call it the TXL-PBC dataset. We used the X-AnyLabeling tool~\cite{XAnyLabeling} to annotate all the datasets. 
Specifically, this study shows below:
\begin{itemize}
    \item We manually selected 160 high-quality images from the BCCD dataset. In addition, we incorporated 100 images from the BCDD dataset available on the HuggingFace open-source platform. Furthermore, we selected 500 cell images each from the PBC and Raabin-WBC datasets. In total, our TXL-PBC dataset consists of 1,260 cell image samples.
    \item We used YOLOv8n to semi-automatically label the TXL-PBC dataset by first manually annotating 250 samples that were selected from the four datasets. These manually labeled samples were used to train the model, which was then used to label the remaining data, followed by a comprehensive manual review\cite{c7,Jocher_Ultralytics_YOLO_2023,yolov8}. All samples were labeled as 'WBC', 'RBC', and 'Platelet'.
    \item We divided the new dataset into a training set (train: 882), a validation set (val: 252), and a test set (test: 126), following a 7:2:1 ratio of the total 1,260 samples.
\end{itemize}

\hspace{2em}Finally, we conducted a comprehensive data visualization analysis to illustrate the distribution and diversity of our TXL-PBC dataset. The results demonstrate that the dataset covers a wide range of cell sizes and types, reflecting its high diversity and annotation quality. In addition, we selected YOLOv5s\cite{yolov5}, YOLOv8s\cite{yolov8}, YOLOv11s\cite{yolov11}, SSD300\cite{SSD}, Faster R-CNN\cite{FasterR-CNN}, and RetinaNet\cite{RetinaNet} as baseline models to train and evaluate the TXL-PBC dataset. The training results showed that TXL-PBC exhibited excellent performance across different detection models.

\section{Material and Methods}\label{sec2}
\subsection{Data Set Selection}
This paper mainly includes four cell image datasets: Blood Cell Count and Detection (BCCD), Blood Cell Detection Dataset (BCDD), Peripheral Blood Cells (PBC), and Raabin White Blood Cell (Raabin-WBC) datasets. Each of these datasets brings unique characteristics and diversity to the integrated TXL-PBC dataset. The details of these datasets are as follows:
\subsubsection{BCCD datasets}
\begin{figure}[htbp]
\centering
    \includegraphics[width=0.7\textwidth]{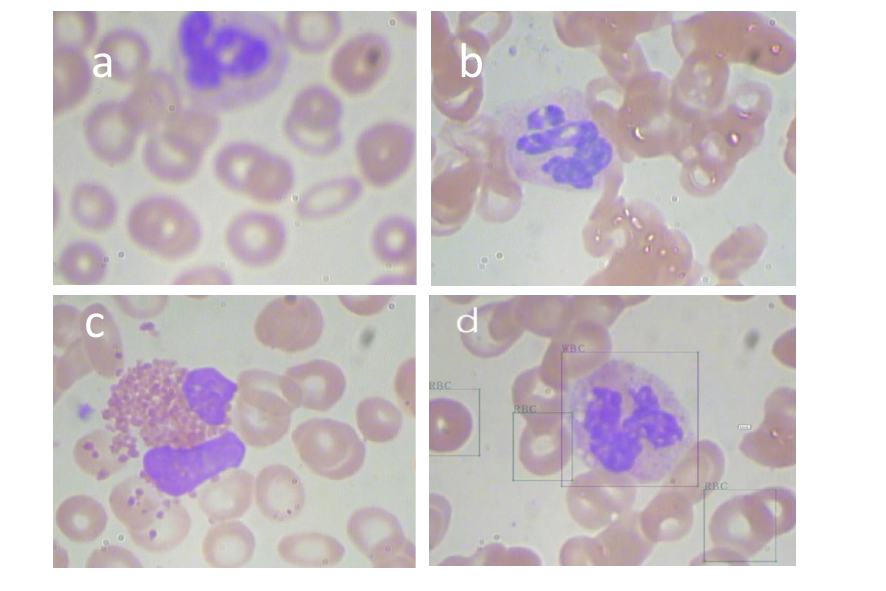}
    \caption{Examples of image quality issues and annotation problems in the BCCD dataset. (a) Out-of-focus or blurry image; (b) Image with severe overlap of red blood cells, making boundaries indistinguishable; (c) Image with damaged white blood cells; (d) Example of a missed annotation problem in the BCCD dataset, where some cells are not labeled.}
    \label{fig1}
\end{figure}

\textbf{Blood Cell Count and Detection} (BCCD) dataset has 364 images, which are classified as “RBC”, “WBC”, and “Platelets”, with a resolution of 640 × 480. This dataset was published on the GitHub website. We manually reviewed all 364 images in the BCCD dataset and performed a detailed cleaning process based on strict criteria. The detailed list of selected and excluded images, including filenames for each category, is available in BCCD selection file, which is provided with the dataset. The quality issues of the images can be clearly observed from Figure~\ref{fig1} (a), (b), and (c). Specifically, images were excluded for the following reasons: (a) out-of-focus or blurry images, which affected the visibility of cell structures (25 images);  (b) severe overlap of red blood cells, which made it impossible to distinguish individual cell boundaries for annotation (172 images); (c) damaged white blood cells, where the nuclei were dispersed outside the cell, making it difficult to identify platelets (7 images). In total, 204 low-quality images were removed, and 160 high-quality images were retained for further annotation and analysis. We have also clearly presented our BCCD Dataset Cleaning and Filtering Results in Table \ref{tab:cleaning_results}. Furthermore, (d) we also found that there are a large number of missed annotations for red blood cells in the dataset, as illustrated in Figure~\ref{fig1} (d).

\begin{table}[H]
\centering

\begin{tabular}{|l|c|}
\hline
\multicolumn{1}{|l|}{\textbf{Filtering Reason}} & \textbf{Number of Images Excluded} \\
\hline
Severe overlap of red blood cells & 172 \\
Out-of-focus or blurry images     & 25  \\
Damaged white blood cells         & 7   \\
\hline
\multicolumn{1}{|l|}{\textbf{Total excluded}} & 204 \\
\multicolumn{1}{|l|}{\textbf{Total retained}} & 160 \\
\hline
\end{tabular}
\caption{Summary of BCCD Dataset Cleaning and Filtering Results}
\label{tab:cleaning_results}
\end{table}

\begin{figure}[htbp]
\centering
    \includegraphics[width=0.6\textwidth]{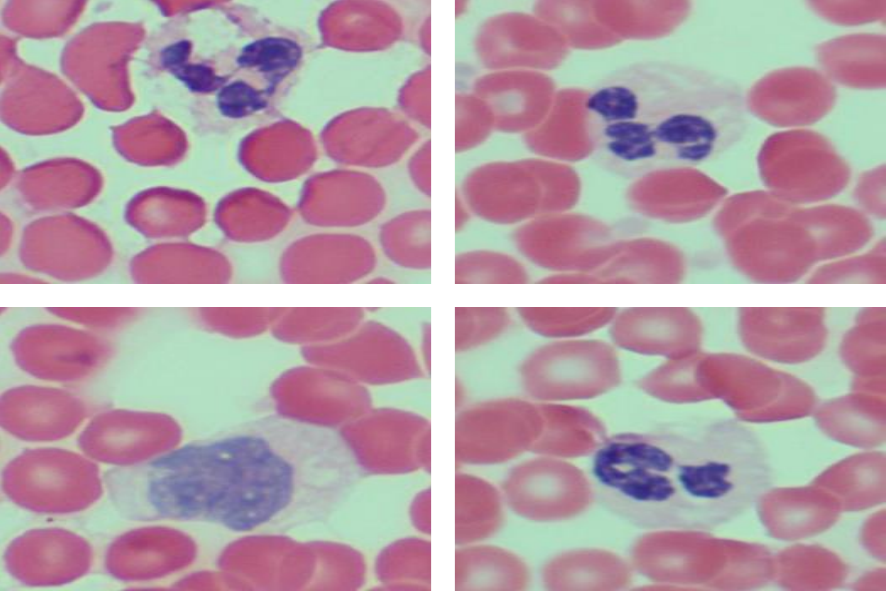}
    \caption{Representative images from the Blood Cell Detection Dataset (BCDD), showing typical peripheral blood smear samples and annotation targets, including red blood cells (RBCs) and white blood cells (WBCs) under high-magnification light microscopy.}
    \label{figa11}
\end{figure}

\textbf{Blood Cell Detection Dataset} is composed of 100 annotated peripheral blood smear images, each captured using a high-magnification light microscope at a resolution of 256$\times$256 pixels in RGB format. The dataset includes 2,237 labeled red blood cells (RBC) and 103 labeled white blood cells (WBC). All images are stored in PNG format within the images folder, and annotation details, including cell locations and labels, are provided in the annotations.csv file. The total dataset size is about 14MB. The dataset is openly available on the HuggingFace website. For our TXL-PBC dataset, we only utilized the images from this dataset and did not use the original annotations, as we aimed to maintain consistency by applying our unified YOLO label format across all datasets. Representative images from the Blood Cell Detection Dataset can be seen in Figure \ref{figa11}.

\begin{figure}[H]
        \centering
        \includegraphics[width=0.6\textwidth]{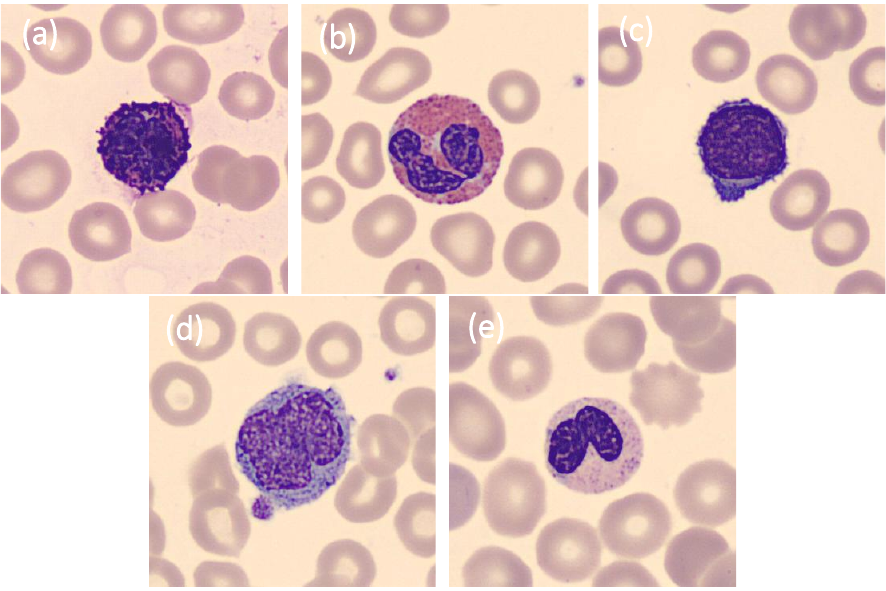}
        \caption{(a) Basophils, (b) Eosinophils, (c) Lymphocytes, (d) Monocytes, (e) Neutrophils.}\label{fig2}
    \end{figure}

\textbf{The Peripheral Blood Cell} (PBC) dataset consists of 17,092 images. These images are further organized into the following eight groups: neutrophils, eosinophils, basophils, lymphocytes, monocytes, immature granulocytes (including promyelocytes, myelocytes, and metamyelocytes), erythroblasts, and platelets or thrombocytes. Each image is 360 x 363 pixels in size and is in JPG format, annotated by expert clinical pathologists. This dataset focuses on images of peripheral blood cells. For our newly introduced dataset, we have selected 500 samples from five types of white blood cells separately in the PBC dataset. Representative images from the PBC dataset can be seen in Figure \ref{fig2}.

\begin{figure}[H]
        \centering
        \includegraphics[width=0.6\textwidth]{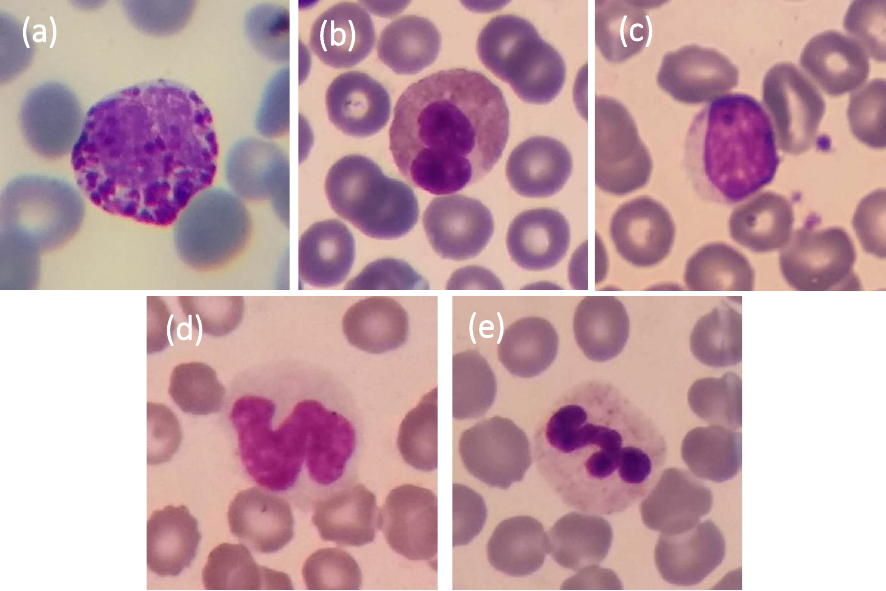}
        \caption{(a) Basophils, (b) Eosinophils, (c) Lymphocytes, (d) Monocytes, (e) Neutrophils.}\label{fig4}
    \end{figure}

\textbf{Raabin White Blood Cell} (WBC) dataset which consisted of 14514 WBC images across five classes 301 basophils, 795 monocytes, 1066 eosinophils, 8891 neutrophils, and 3461 lymphocytes at resolutions of 575 x 575. The data set mainly focuses on the classification of white blood cells. We selected 100 samples form each type of white blood cells and introduced 500 samples into our new data set. And the total As shown in Figure\ref{fig4}.

\subsection{Data Integration}
\begin{figure}[htbp]
    \centering
    \includegraphics[width=0.6\textwidth]{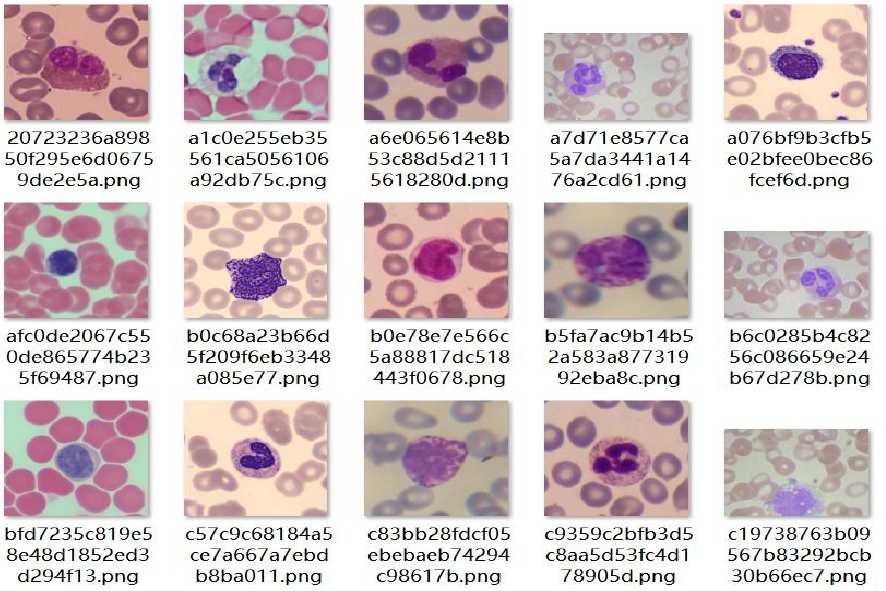}
    \caption{ Overview of the TXL-PBC dataset after integration and randomization. The figure shows a sample of images from the combined dataset, demonstrating the diversity and random naming of the images.}\label{fig5}
\end{figure}

In this paper, our goal is to build a new dataset that is balanced, diverse, and of high quality. Therefore, we combined all the datasets, resulting in a total of 1,260 samples: 160 samples from BCCD, 100 samples from BCDD, and 500 samples each from the PBC and Raabin-WBC datasets. During the integration process, we randomly shuffled and renamed the images to ensure a high level of randomness and diversity in the samples. An overview is shown in Figure~\ref{fig5}.

In addition, we provide a metadata file for the TXL-PBC dataset, which maps each image filename to its original source dataset and includes the corresponding source URL. This metadata allows users to trace the provenance of each image and facilitates reproducibility and further data integration.

\subsection{Data Annotation}
\begin{figure}[htbp]
    \centering
    \includegraphics[width=0.6\textwidth]{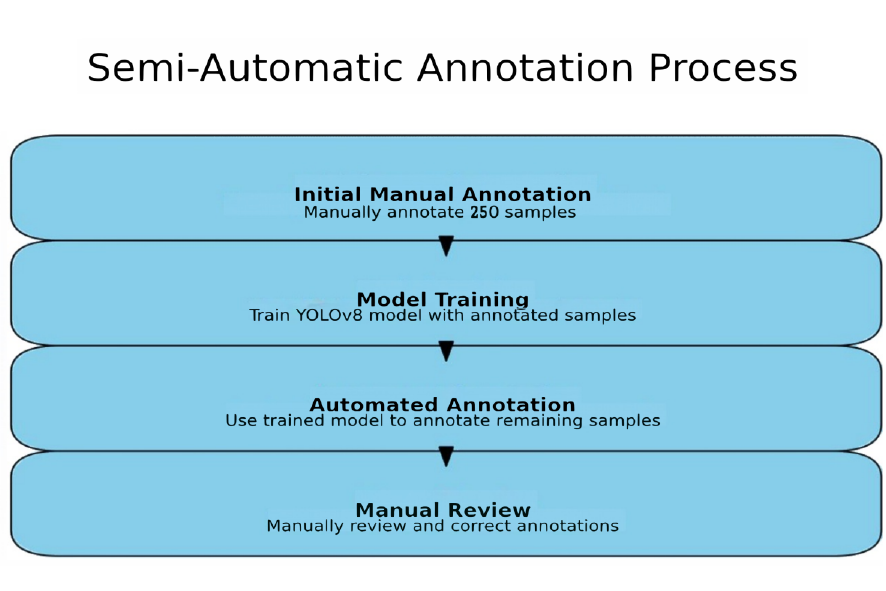}
    \caption{Workflow of the semi-automatic annotation process. Initial manual annotations were used to train a YOLOv8n model, which then automatically labeled the remaining images. All annotations were subsequently reviewed and corrected manually to ensure high accuracy.}\label{fig3}
\end{figure}

To improve the quality of the datasets and save manual annotation time, we adopted a semi-automatic annotation method using the YOLOv8n model. The complete workflow is illustrated in Figure \ref{fig3}. This semi-automatic labeling method greatly improves annotation efficiency. The specific method is as follows.

This semi-automatic labeling method greatly improves the efficiency of labeling. Combined with manual screening, it ensures the accuracy and consistency of the labels. We applied this approach to all datasets, resulting in high-quality annotated data. This provide a reliable basis for subsequent data integration and analysis.
\subsubsection{Initial Manual Annotation}
To train the YOLOv8n model for semi-automatic annotation, we manually selected 250 images from the four datasets for initial manual labeling: 100 images from the PBC dataset, 100 images from the Raabin-WBC dataset, 30 images from the BCCD dataset, and 20 images from the BCDD dataset. This distribution strategy was designed to ensure that the model could learn the distinctive characteristics of each dataset equally, thereby enhancing its robustness and generalization capability across different data sources. During the manual labeling process, we placed particular emphasis on the precision of the cell boundary boxes and the accurate identification of the three types of cells in the images.

For manual annotation, we used the open-source tool X-AnyLabeling\cite{XAnyLabeling}. We chose this tool because X-AnyLabeling is a versatile graphical image annotation tool that supports multiple annotation formats, including the YOLO format used in our project. Using its intuitive interface, annotators can easily open image folders, select the desired annotation format, and draw bounding boxes around each cell of interest by clicking and dragging. Each bounding box can be assigned the correct label (e.g., RBC, WBC, Platelet), and the annotation files are saved in the YOLO format in the same directory as the images. X-AnyLabeling also provides convenient keyboard shortcuts to improve annotation efficiency and ensure consistency across different annotators.
We consistently used the X-AnyLabeling tool following our standardized protocol, which allowed us to accurately draw bounding boxes and provided the necessary functionality to ensure the consistency and accuracy of the labeled samples.

To ensure consistency and accuracy across all manual annotations, we developed a comprehensive manual annotation protocol that includes detailed guidelines for annotation tool usage, clear definitions of annotation targets, standardized annotation procedures, quality control measures, and suggestions for handling common annotation challenges. This protocol was strictly followed by all annotators throughout the manual labeling process to maintain high annotation quality and inter-annotator consistency. The complete manual annotation protocol is available as supplementary material and can also be accessed through our GitHub repository and Figshare website.

\subsubsection{Model Training}
YOLO (You Only Look Once) is a widely used single-stage object detection framework known for its real-time performance and high accuracy. YOLOv8n is the nano version of the latest YOLOv8 series, designed to be both lightweight and powerful. YOLOv8n was chosen among many potential models due to its excellent balance between speed and accuracy. It is particularly suitable for tasks that require quick inference times without sacrificing detection precision. The nano version (YOLOv8n) is especially well-suited for our semi-automatic annotation task as it provides fast training and inference while maintaining sufficient accuracy for blood cell detection.

\begin{figure}[htbp]
    \centering
    \includegraphics[width=0.6\textwidth]{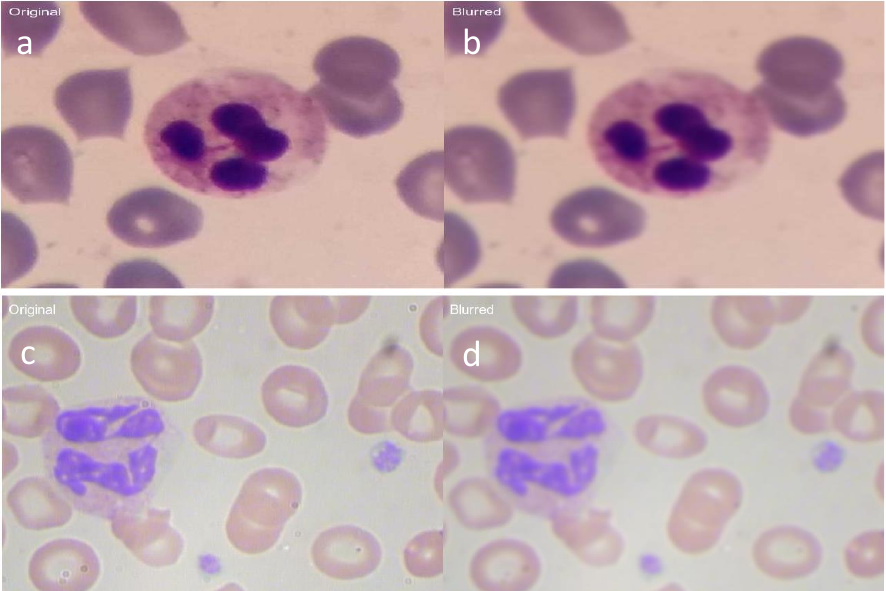}
    \caption{Comparison between original images and Gaussian blurred versions used for data augmentation. (a) Original blood cell image; (b) Corresponding Gaussian blurred version; (c) Another original blood cell image; (d) Its corresponding Gaussian blurred version. This augmentation technique effectively doubles our training data while preserving essential cell characteristics for improved model robustness.}
    \label{fig_gaussian}
\end{figure}

Since we only used 250 manually annotated samples for initial training, we applied Gaussian blur augmentation to increase our training data to 500 samples, thereby improving the model's robustness capability. We used Gaussian smoothing filters to create augmented versions of our original samples\cite{Gaussianblur,Gaussian}. This augmentation technique helps the model focus on key features instead of noise and improves its ability to perform well under different imaging conditions and handle images of varying quality\cite{Gaussian2,Gaussian3}. The Gaussian blur augmentation effectively doubles our training data while maintaining the essential characteristics of blood cells. Figure \ref{fig_gaussian} shows examples of original images and their corresponding Gaussian blurred versions used for data augmentation.

The formula for applying a Gaussian filter to an image \( I(x, y) \) is:

\[
I'(x, y) = \sum_{i=-k}^{k} \sum_{j=-k}^{k} I(x+i, y+j) \cdot G(i, j)
\]

where \( G(i, j) \) is the Gaussian kernel, and \( k \) is the filter window size.

To maximize the performance of YOLOv8n, we trained the model using the following parameters: 100 epochs, a batch size of 32, an image size of 512×512, zero workers, and the AdamW optimizer. These settings were chosen to balance training efficiency and model accuracy, ensuring that the model could effectively learn from the augmented data without overfitting or underfitting. The relatively small batch size and moderate image resolution were selected to accommodate the limited training data while maintaining sufficient detail for accurate cell detection.

\subsubsection{Automated Annotation}
In the automated annotation process, we used a trained YOLOv8n model to predict and label unannotated data, filtering predictions by confidence to improve accuracy. For complex edge cases, such as overlapping objects and blurry boundaries, we applied Non-Maximum Suppression (NMS)\cite{Efficient_non-maximum_suppression,NMS,NMS1,NMS2}. We also flagged low-confidence annotations for manual review to ensure data reliability and precision.

\textbf{Confidence Filtering}:
   \[
   B_{\text{filtered}} = \{ B_i \mid s_i > t_c \}
   \]
   where \( B_i \) is the bounding box, \( s_i \) is the corresponding confidence score, and \( t_c \) is the set confidence threshold. We set the confidence threshold at 0.5, meaning that only when the model's predicted confidence for a target exceeds 50\% will the annotation be retained.

\textbf{Non-Maximum Suppression (NMS)}:
   \[
   B_{\text{NMS}} = \{ B_i \mid s_i > t, \text{IoU}(B_i, B_j) < t \, \forall \, B_j \neq B_i \}
   \]
   where \( \text{IoU}(B_i, B_j) \) is the Intersection over Union between two bounding boxes, and \( t \) is the IoU threshold.

\begin{figure}[htbp]
    \centering
    \includegraphics[width=0.6\textwidth]{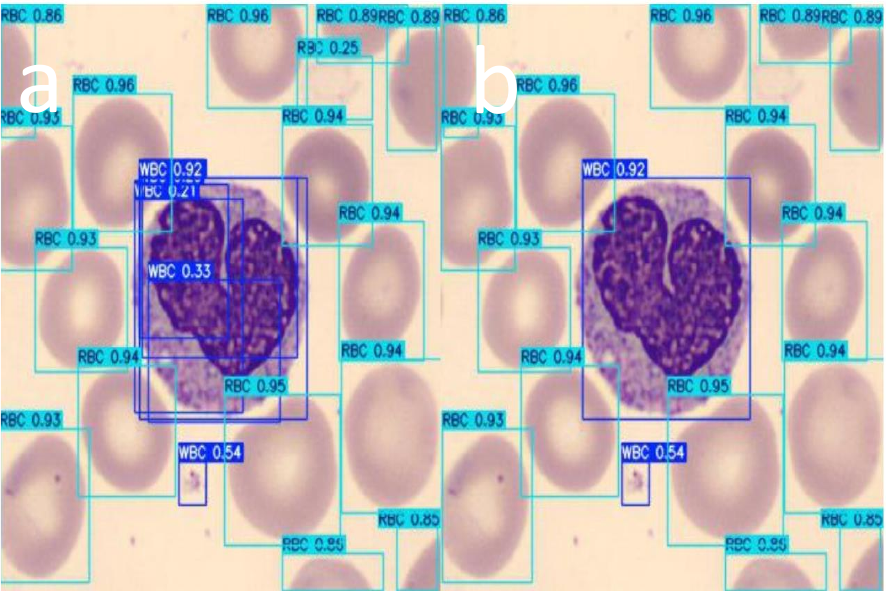}
    \caption{Examples of confidence filtering in the automated annotation process. (a) Example with confidence threshold below 0.5, showing multiple overlapping bounding boxes with low confidence scores for white blood cells; (b) After applying confidence filtering with threshold above 0.5, only accurate and reliable bounding boxes are retained for white blood cells. This demonstrates the effectiveness of confidence filtering in reducing false positives and improving annotation quality.}\label{fig111}
\end{figure}

The final results generated by the trained model are shown in Figure \ref{fig111}. These images demonstrate the effectiveness of confidence filtering in the automated annotation process. As shown in (a), when the confidence threshold is below 0.5, multiple overlapping bounding boxes with low confidence scores appear for white blood cells, leading to redundant and inaccurate annotations. In contrast, (b) shows that after applying confidence filtering with a threshold above 0.5, only accurate and reliable bounding boxes are retained for white blood cells. This comparison clearly illustrates how confidence filtering effectively reduces false positives and improves annotation quality in our semi-automatic labeling approach.

\subsubsection{Manual Review}
To ensure the reliability and accuracy of our data, we conducted a comprehensive manual review of all samples automatically annotated by the models, with particular attention given to annotations with a confidence level below 0.5. To further validate the quality of our annotations, we randomly selected 200 samples and invited local pathologists to conduct an independent review. The pathologists examined these samples, and the results showed that our annotations were in full alignment with the pathologists' professional standards. This not only confirmed the reliability of our annotation methods but also provided a solid foundation for the subsequent application of the data. Through this multi-layered review process and the involvement of external experts, we were able to ensure the high quality of our data, thereby laying a strong foundation for future research and analysis.
\subsection{Dataset Splitting}
We divided the data set into a training set, a validation set, and a test set in a 7:2:1 ratio. The training set consists of 1,008 samples for machine learning model training. The validation set has 288 samples. It is used to adjust model parameters during training and to prevent over-fitting. The test set contains 144 samples that are used to finally evaluate the performance of the model. We named the new dataset TXL-PBC. We aim to provide a high-quality sample set for cell research and machine learning models.

\section{Results}\label{sec3}
\subsection{Sample Comparison} 
\begin{figure}[htbp]
    \centering
    \includegraphics[width=0.6\textwidth]{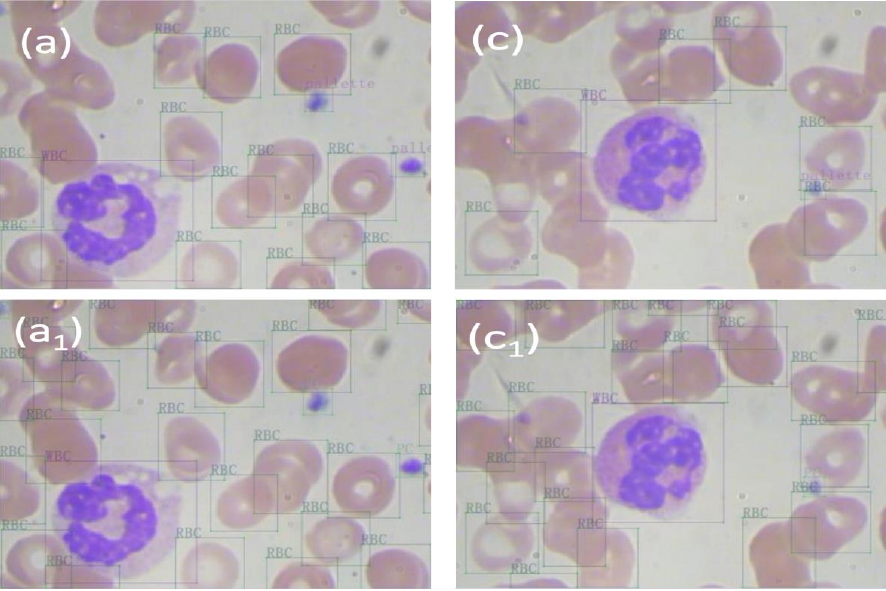}
    \caption{Comparison of annotation quality between the original BCCD dataset and the TXL-PBC dataset. (a) and (c): Original images from BCCD dataset showing missing or incorrect labels for red blood cells. (a$_1$) and (c$_1$): Corresponding images from the TXL-PBC dataset where all blood cells are accurately labeled. This comparison demonstrates that the TXL-PBC dataset successfully corrected numerous labeling errors and missed annotations present in the original dataset.}\label{fig7}
\end{figure}

To demonstrate the significant improvement in the annotation quality of the TXL-PBC dataset, we randomly selected three images from the BCCD samples in our new dataset and compared them with the corresponding images from the original BCCD dataset. (a, c) represents the original BCCD dataset, and (a$_1$, c$_1$) represents the same images from our TXL-PBC dataset. As shown in Figure \ref{fig7}, it is evident that many RBCs are missing or incorrectly labeled in the original dataset, whereas they are all accurately labeled in our TXL-PBC dataset. This comparison clearly demonstrates that our dataset has successfully corrected numerous labeling errors and missed annotations that existed in the original datasets. 

\begin{figure}[htbp]
    \centering
    \subfigure[Width-Height Distribution of Bounding Boxes]{\includegraphics[width=0.4\textwidth]{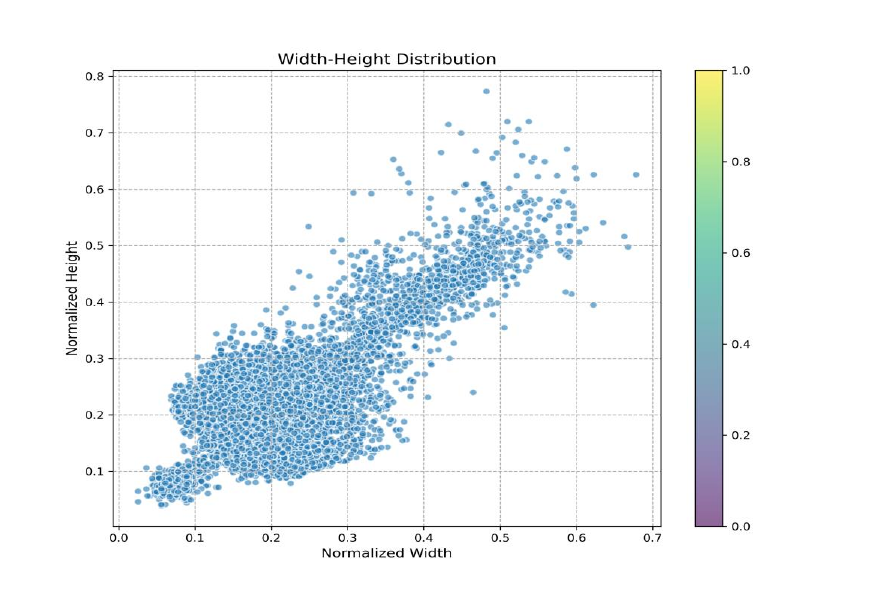}}
    \hfill
    \subfigure[Normalized Coordinate Distribution]{\includegraphics[width=0.4\textwidth]{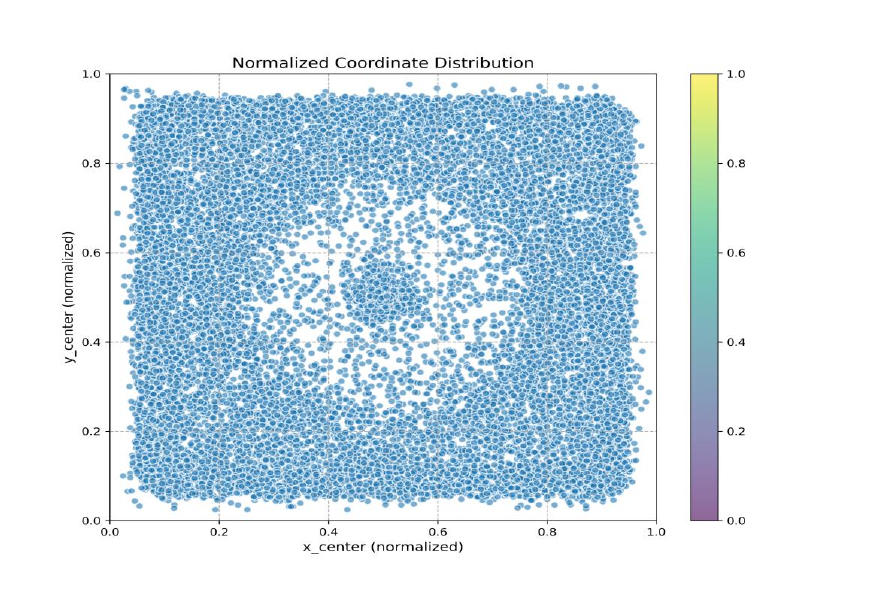}}
    \hfill
    \subfigure[Class distribution of annotated cells]{\includegraphics[width=0.4\textwidth]{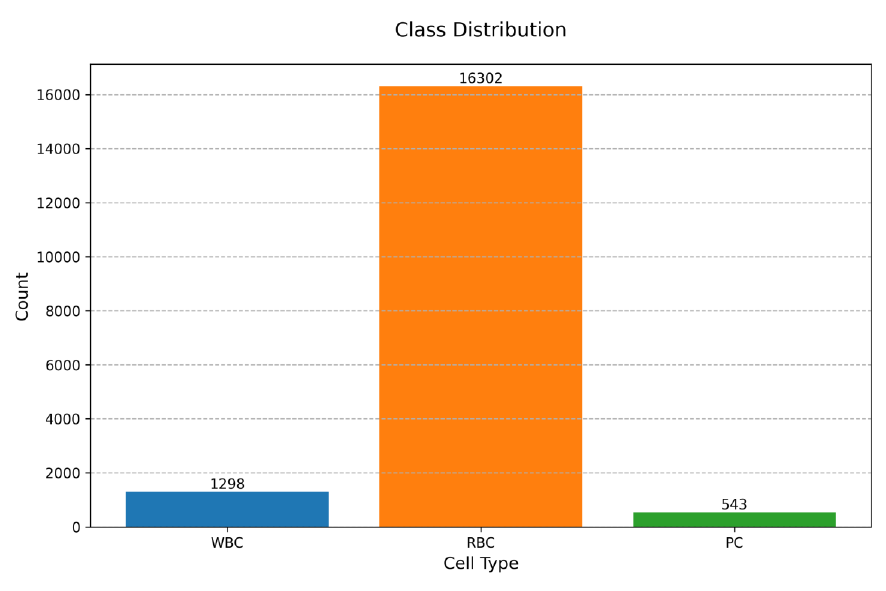}}
    \caption{Statistical analysis of the TXL-PBC dataset. (a) Width-Height Distribution of Bounding Boxes: the scatter plot shows the diversity of cell sizes. (b) Normalized Coordinate Distribution: bounding box centers are evenly distributed across the image area. (c) Class distribution of annotated cells: bar chart showing the number of RBC, WBC, and platelets (PC) annotations.}
    \label{fig6}
\end{figure}

\subsection{ Dataset Statistical Analysis} 
We conducted a comprehensive statistical analysis of the TXL-PBC dataset to evaluate its diversity and representativeness. First, the width and height distribution of all bounding boxes (Figure~\ref{fig6}a) shows a wide range of normalized values, reflecting the diversity of cell sizes present in the dataset. Second, the scatter plot of normalized bounding box centers (Figure~\ref{fig6}b) demonstrates that the cells are evenly distributed across the entire image area, with no obvious clustering or missing regions. This uniform spatial distribution helps improve the generalization ability of object detection models and prevents overfitting to specific regions. Finally, the class distribution of annotated cells (Figure~\ref{fig6}c) reveals that red blood cells (RBC) account for the majority of annotations (16,302), while white blood cells (WBC) and platelets (PC) have 1,298 and 543 annotations, respectively. This distribution is consistent with the natural proportions of different cell types in peripheral blood, indicating that the TXL-PBC dataset provides a comprehensive and realistic resource for blood cell detection and classification tasks.

\subsection{Baseline Models and Their Performance Evaluation}
\begin{figure}[htbp]
    \centering
    \includegraphics[width=0.7\textwidth]{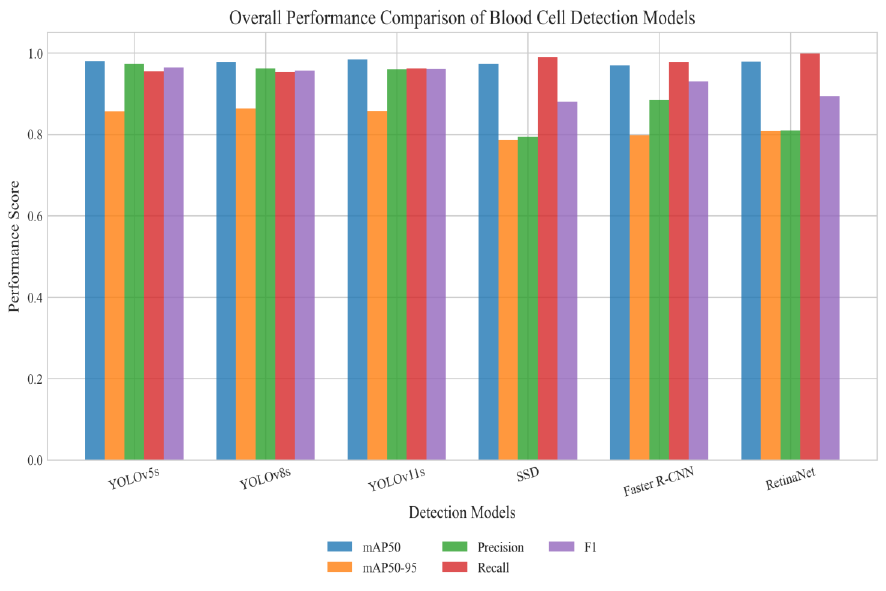}
    \caption{Overall performance comparison of different blood cell detection models on the TXL-PBC dataset. The grouped bar chart shows the mAP50, mAP50-95, precision, recall, and F1-score for each baseline model, providing a clear visual comparison of their overall detection performance.}
    \label{fig:model_performance}
\end{figure}

In this study, we selected six representative object detection models as baselines to comprehensively evaluate the TXL-PBC dataset: YOLOv5s, YOLOv8s, YOLOv11s, SSD300, Faster R-CNN, and RetinaNet. The YOLO series (v5s, v8s, v11s) are widely used single-stage detectors known for their fast inference speed and good balance between accuracy and efficiency. YOLOv5s is a mainstream version commonly adopted in biomedical image analysis, while YOLOv8s and YOLOv11s represent the latest advancements in the YOLO family, allowing us to compare the impact of architectural improvements across versions. SSD300 is a classic single-stage detector with a simple structure and multi-scale feature maps, making it a popular baseline in object detection research. Faster R-CNN is a two-stage detector that achieves high accuracy through its region proposal network, serving as a strong reference for accuracy comparison with single-stage models. RetinaNet is a single-stage detector that introduces focal loss to address class imbalance, combining the advantages of both single- and two-stage approaches. By including these models, we aim to provide a comprehensive and fair benchmark, covering a range of detection paradigms and algorithmic innovations, and to offer valuable references for future research on blood cell detection tasks. The parameters consistently set for all baseline models were: training for 100 epochs, a batch size of 16, an image size of 320x320, zero workers, the AdamW optimizer, caching enabled to improve training efficiency, and an initial learning rate of 0.001. The performance of the models is shown in table \ref{tab:results}. In addition, we visualized the overall results of these six models using grouped bar charts to provide a more intuitive comparison, as shown in Figure~\ref{fig:model_performance}.
\begin{table}[h]
\centering
\begin{tabular}{llccccc}
\hline
\textbf{Model} & \textbf{Type} & \textbf{mAP50} & \textbf{mAP50-95} & \textbf{Precision} & \textbf{Recall} & \textbf{F1} \\
\hline
YOLOv5s      & Overall & 0.980 & 0.857 & 0.973 & 0.955 & 0.964 \\
             & WBC     & 0.994 & 0.899 & \textbf{0.994} & 0.996 & \textbf{0.995} \\
             & RBC     & 0.989 & 0.918 & 0.953 & 0.968 & 0.960 \\
             & Platelets      & 0.956 & 0.755 & 0.971 & 0.900 & 0.937 \\
YOLOv8s      & Overall & \textbf{0.978} & \textbf{0.864} & 0.962 & 0.952 & 0.957 \\
             & WBC     & 0.992 & 0.899 & 0.978 & 0.996 & 0.987 \\
             & RBC     & \textbf{0.989} & \textbf{0.918} & 0.953 & 0.960 & 0.960 \\
             & Platelets      & 0.954 & 0.774 & 0.955 & 0.893 & 0.923 \\
YOLOv11s     & Overall & 0.984 & 0.858 & 0.959 & 0.955 & 0.957 \\
             & WBC     & \textbf{0.994} & 0.895 & 0.981 & 0.996 & 0.988 \\
             & RBC     & 0.990 & 0.916 & \textbf{0.961} & 0.961 & 0.961 \\
             & Platelets      & 0.969 & 0.764 & 0.937 & 0.928 & 0.932 \\
SSD300       & Overall & 0.973 & 0.786 & 0.794 & \textbf{0.989} & 0.881 \\
             & WBC     & 0.990 & 0.838 & 0.959 & 0.992 & 0.977 \\
             & RBC     & 0.975 & 0.820 & 0.802 & \textbf{0.988} & 0.885 \\
             & Platelets      & 0.954 & 0.699 & 0.621 & 0.982 & 0.761 \\
Faster R-CNN & Overall & 0.969 & 0.798 & 0.884 & 0.978 & \textbf{0.929} \\
             & WBC     & 0.989 & 0.816 & 0.981 & 0.992 & \textbf{0.989} \\
             & RBC     & 0.976 & 0.848 & 0.877 & 0.987 & 0.929 \\
             & Platelets      & 0.941 & 0.730 & 0.793 & 0.955 & 0.867 \\
RetinaNet    & Overall & 0.979 & 0.808 & \textbf{0.809} & 0.998 & 0.894 \\
             & WBC     & 0.991 & \textbf{0.855} & 0.931 & \textbf{1.000} & 0.964 \\
             & RBC     & 0.978 & 0.839 & 0.814 & 0.993 & \textbf{0.895} \\
             & Platelets      & \textbf{0.968} & \textbf{0.730} & 0.683 & \textbf{1.000} & \textbf{0.812} \\
\hline
\end{tabular}
\caption{Performance comparison of different models on the TXL-PBC dataset.}
\label{tab:results}
\end{table}

We evaluated the performance of each baseline model on the TXL-PBC dataset, including detailed results for white blood cells (WBC), red blood cells (RBC), and platelets. The YOLO series models (YOLOv5s, YOLOv8s, and YOLOv11s) all achieved high detection accuracy, with overall mAP50 values around 0.98. YOLOv8s achieved the highest overall mAP50 (0.978) and mAP50-95 (0.864), while YOLOv5s showed the highest precision (0.994) for WBC and the highest F1-score (0.995) for WBC. YOLOv11s achieved the highest precision (0.961) for RBC and balanced performance across all metrics. SSD300 demonstrated the highest recall (0.989) overall and for RBC (0.988), but its precision and F1-score were lower compared to the YOLO models. Faster R-CNN achieved the highest overall F1-score (0.929) and the highest F1-score for WBC (0.989), indicating an excellent balance between precision and recall. RetinaNet achieved the highest recall for WBC and platelets (both 1.000), the highest mAP50 (0.968) and mAP50-95 (0.730) for platelets, and the highest F1-score for platelets (0.812), but its overall precision was lower. These results highlight the strengths of different models for specific cell types and metrics. The YOLO series and Faster R-CNN provide a good balance of accuracy and reliability for most categories, while SSD and RetinaNet show trade-offs between recall and precision, especially for platelets. This comprehensive comparison demonstrates the robustness and diversity of the TXL-PBC dataset and provides valuable references for future research and model selection in blood cell detection tasks.

\section{Discussion}\label{sec4}

Our research results show that TXL-PBC dataset not only has significant advantages in the number of labels and boundary box area distribution, but also far exceeds the actual detection effect of existing data sets. This indicates that the data set has great application potential in medical image analysis, which can be used to assist diagnosis, disease research and other fields, and is expected to provide important support for medical image processing and analysis.

\hspace{2em}Although we have improved the sample diversity, balance, and quality of our cellular datasets, our work still has limitations. Firstly, the sample diversity and quantity of the TXL-PBC dataset need further expansion. Additionally, despite our strict screening process when labeling cell samples, the consistency of labeling may be affected by subjective differences among different annotators.

\hspace{2em}Therefore, our work will focus on expand the size and diversity of the dataset to enhance the model's generalization ability. Simultaneously, to improve the accuracy and efficiency of annotation, we will explore more effective annotation tools and methods. With these improvements, we aim to provide stronger technical support for cell object detection tasks, automatic cell image labeling, and machine learning models.

\section{Conclusion}\label{sec5}

In this paper,we introduced the TXL-PBC dataset, a high-quality and diverse peripheral blood cell image collection created by integrating and re-annotating four public datasets: BCCD, BCDD, PBC, and Raabin-WBC. Through rigorous sample selection, semi-automatic annotation with YOLOv8n, and manual review, we ensured accurate and consistent labeling. The final dataset contains 1,260 images and 18,143 bounding boxes for RBC, WBC, and platelets. We validated TXL-PBC by benchmarking six mainstream object detection models, demonstrating strong and robust detection performance across all models. TXL-PBC addresses annotation errors and data scarcity in previous datasets, and is openly available on Figshare and GitHub as a valuable resource for blood cell detection research.

\section*{Availability of Data and Materials}
We have published the TXL-PBC dataset on Github (https://github.com/lugan113/TXL-PBC\_Dataset). We hope that more researchers will use this dataset for further studies. We hope that through our work, we can promote the development of cell detection technology and promote the progress of related medical image detection.
\section*{Acknowledgments}
Special thanks to all the people and institutions who provided Blood cell datasets and technical support in this study, without their help, this study could not have been successfully completed.

\end{document}